# An Explainable Model for EEG Seizure Detection based on Connectivity Features

**Mohammad Mansour, Fouad Khnaisser, and Hmayag Partamian**

American University of Beirut

{mgm35, fgk06, hkp00}@mail.aub.edu

*Abstract—* **Epilepsy which is characterized by seizures is studied using EEG signals by recording the electrical activity of the brain. Different types of communication between different parts of the brain are characterized by many state of the art connectivity measures which can be directed and undirected. We propose to employ a set of undirected (spectral matrix, the inverse of the spectral matrix, coherence, partial coherence, and phase-locking value) and directed features (directed coherence, the partial directed coherence) to learn a deep neural network that detects whether a particular data window belongs to a seizure or not, which is a new approach to standard seizure classification. Taking our data as a sequence of ten sub-windows, we aim at designing an optimal deep learning model using attention, CNN, BiLstm, and fully connected layers. We also compute the relevance using the weights of the learned model based on the activation values of the receptive fields at a particular layer. Our best model architecture resulted in $97.03\%$ accuracy using balanced MIT-BIH data subset. Also, we were able to explain the relevance of each feature across all patients. We were able to experimentally validate some of the scientific facts concerning seizures by studying the impact of the contributions of the activations on the decision.**

## I. Introduction

Epilepsy is a neurological disorder that affects around 50 million people of all ages worldwide [1]. It is characterized by the frequent and repetitive occurrence of seizures that disrupt normal function and affect the quality of life of the patient.
Neuroimaging techniques acquired from epilepsy patients showed evidence of different structural and functional irregularities. Synchronous discharges of electrical activity across different parts of the brain during seizures were captured using electroencephalogram (EEG) data [2]. To extract meaningful information, a plethora of measures have been designed to characterize the changes in the EEG signals during epilepsy that can be classified into two types. Univariate metrics measure the information in a window of one time series and can be classified as temporal, spectral, and entropy-based measures [3][4][5]. On the other hand, multivariate connectivity metrics [6][7][8], characterize information between multiple time series and can be directed (effective) or undirected (functional). For example, the phase-locking value (PLV) is an undirected measure since it quantifies the synchrony between a pair of signals [9]. Granger Causality (GC) calculates the amount of information transfer from one channel to the other which makes it a directed measure [10][11][12]. Cross-frequency coupling (CFC) methods have also been used in characterizing information between different brain regions by studying the interaction between oscillations across different frequency bands such as phase-amplitude coupling (PAM) [13].

Ictal episodes are those that exhibit seizure activity and will have different connectivity values when compared to non-ictal periods [14]. Also, seizure onset zones usually get isolated in their activity before the seizure starts which can be captured using a coherence connectivity matrix [15]. Electrophysiological research also reports that functional connectivity analysis allows localization of seizure onset zones (SOZ), whose exact location helps increase surgery success rates [3]. Besides, variations in phase-locking value (PLV) can characterize the synchronous activity between different parts of the brain during a seizure and non-seizure episodes [9].

In the last decade, advances in technology have made machine learning and big data analysis available and many seizure detection algorithms were developed using diverse machine learning algorithms. Support vector machines (SVM) is one of the common techniques used to learn classification algorithms for seizure detection [16][17][3]. Usually, many classical tailored features are fed to the SVM classifiers. Therefore, employing unnecessary features may hinder both the performance and the speed of analysis [18]. Feature selection techniques were proposed to reduce the feature set since machine learning algorithms often perform better when relevant features are selected and used for learning [19]. An alternative branch of machine learning is deep learning (DL) which studies multiple-layer architectures of neural networks that can learn discriminating features and a classifier simultaneously [20]. The raw EEG data, snapshot images, different univariate and bivariate measures were used as input to deep classifiers to detect seizures and non-seizure episodes during epilepsy [21][22][23]. However, deep networks are perceived as "black box" techniques since the role of the different layers and the overall internal functioning are unknown. A major issue arises because scientists need to understand why such a decision was made. Can such models be trusted without knowing why they fail and why they succeed? Explainable artificial intelligence (XAI)[24] techniques try to derive explanations from the parameters of the deep network to infer knowledge and build explainable features [13].

In epilepsy analysis, seizure detection is an important problem, however, for doctors to be able to diagnose epilepsy, they need deeper information about the interactions of the brain regions such as the localization of SOZ which can help identify a resection area for surgery. Experimental studies have shown the phase-amplitude coupling [13] and the power in the high-frequency band (>100Hz) [25] increase in the seizure onset zone and during the seizures. The SOZ also gets disconnected from the rest of the brain regions that can be inferred using undirected connectivity measures such as PLV [26]. Another study shows that the phase-lag index (PLI) connectivity in the θ (4-8 Hz) band can be linked to tumor-related epilepsy [27]. Explainability during seizure analysis were also addressed to find the feature relevance at different frequency bands [28] and extract information from the learned weights to derive topographic brain maps [28][29].

In this study, we selected a subset of the different connectivity measures to characterize the brain signals from different perspectives. spectral matrix (SM), the inverse of the spectral matrix (IS), coherence (COH), partial coherence (PC), and phase-locking value (PLV) as undirected measures, and directed coherence (DC), and the partial directed coherence (PDC) as directed features. We feed the chosen features to learn a deep model that not only classifies seizures but also computes how much each of the features has participated in the decision made by the detector using the deep learning model parameters. The weights of the model provide explainability at the level of the features. This is of key importance for epilepsy analysis since it provides the user with additional valuable information. Also, seizure events differ from one user to the other and the model will be generally trained using different patient data that have different seizure dynamics. The types of epilepsy in the training set can also be different and may exhibit different connectivity values [30]. Therefore, when new data is tested, the proposed model will detect whether a specific interval of EEG is a part of a seizure or not and also outputs the percentage of the impact of each of the employed connectivity measures on the model's decision.

We summarize our contributions as follows:
- Unlike classical methods where they use a single window, our model considers a 20-second window with ten 2 second sub-windows to characterize interdependencies between these windows through time and better describe seizure and non-seizure intervals of the EEG data.
- The model is designed to perform seizure detection of the 20-second data using attention, convolutional neural networks (CNN), fully connected layers (FC), and bidirectional long short term memory (BiLstm) methodologies.
- In [37], different fusion methods were employed while building the architecture of the models. We also employed different fusion methods and compared them to understand relationships between features and the output
- Finally, we infer from the weights of the network the relevance of the connectivity measures on the decision made by the detector. We study the impact of the features on the output generally across all patients as well as across patients. To the best of our knowledge, no other work has studied explainability with connectivity analysis during seizure classification.

TABLE I
COMPARABLE ANALYSIS

| Ref. | Task | Feature Selection | Deep Architecture | Acc. | XAI |
|---|---|---|---|---|---|
| [23] | Seizure Detection | Raw, Spectral, temporal, EEG snapshot, spectrogram | FCNN, RNN, DNN | 99.7% | No |
| [28] | Seizure Detection | Raw Data | CNN | 98.05% | Yes |
| [29] | Seizure Detection | Raw Data | CNN, Attention, FCNN | - | Yes |
| [31] | Seizure Prediction | Raw data | AE+ BiLstm | 99.66% | No |
| [32] | Seizure Detection | Directed Connectivity and Graph Metrics | DNN | 99.43% | No |
| [37] | Video Detection | Trajectory Features | DNN | 93.33% | Yes |
| [39] | Schizophrenia Detection | Mixed Connectivity | CNN | 91.69% | No |
| [41] | Seizure Detection | Raw Data | CNN, BiLstm | 98.89% | No |
| Proposed Method | Seizure Detection | 7 Connectivity Measures | CNN, BiLstm, Attention, FCNN | 95.54% | Yes |

The rest of the paper is organized as follows. In section II, we present all the related work for this study. In section III, we explain our methodology providing the different designs and workflows of the proposed method. In section IV, a series of experiments are conducted to evaluate the performance of our design. In section V, we discuss our findings, its limitations, and propose possible future extensions. We finally conclude in section VI, providing a summary of our work.

## II. RELATED WORK

Two tasks arise while working with epileptic data, classifying the sample as ictal or non-ictal or classifying the window as preictal (the period that precedes an ictal phase) or inter-ictal (period of normal activity). Even though we will tackle the ictal/non-ictal classification problem, studying the preictal/inter-ictal problem is useful as it gives us intuition and inspiration to develop our architecture. The state of the art for preictal and inter-ictal classification is produced by Daoud et al. [31] where they describe a methodology to train, using the Raw Multichannel EEG data, a deep convolutional autoencoder (DCAE) and uses the latent space representation i.e. (output of the pre-trained encoder) of each recording as input to a BiLstm network that classifies the example into preictal and inter-ictal. The Dataset used is the CHB-MIT EEG dataset recorded at Children's Hospital Boston and is publicly available. To narrow down the channels considered (23 total channels), an iterative algorithm was used to select the channels. The algorithm calculates the product between variance and entropy for each channel and iteratively trains the model on bigger windows until all the channels are considered to select the best combination of performance accuracy and computational cost. Through this approach, an accuracy of 99.66% is achieved.

As for ictal/non-ictal prediction, Akbarian et al. [32] use effective brain connectivity, Directed Transfer Function (DTF), Directed Coherence (DC), and Generalized Partial Directed Coherence (GPDC) in various frequency bands, to measure the relation between brain regions. They extracted features from each of these measures using graph theory. They fed each set of newly extracted features to an autoencoder (AE) for feature reduction then used a softmax on the output of the encoder, obtaining 3 classifiers. The decision of ictal/non-ictal is then inferred through majority voting. Through this method and using the CHB-MIT dataset, they achieved a 99.43% accuracy. Table 1 presents an overview of different architectures and their respective task-specific accuracies. We also include the features used by each study. When using different features we need to combine them to feed them to the network, different types of combinations/fusions exist such as concatenating them at the input layer, or feeding them through separate networks and concatenating their respective outputs and use them as input to the final output layer. Fusion methods were applied in EEG with the Schizophrenia detection problem where they showcase the effect of varying fusion mechanisms on the performance of the deep network [39]. Another approach for epilepsy detection uses the raw EEG data as input to a bidirectional recurrent network that uses the past window knowledge to predict the next window's label. The method can discriminate normal-ictal and normal-ictal-interictal EEG signals accurately [41].

New techniques are currently developing and shaping into a field of study called explainable AI (XAI) [24] where researchers try to make use of the learned blocks of information inside the deep learning models. A modified deep learning model will also learn explainable features while training. The first efforts employed the deep CNN using deconvolution methods to explain the feature maps. Parts of the image that activated certain neurons were marked during the process [33]. Another XAI algorithm, LIME (Local Interpretable Model-Agnostic Explanations), finds for every test sample the relevance of a particular learned feature for a specific output using a local approximation with a linear sparse model [34]. Class activation mapping (CAM) is another method for saliency map generation primarily used for object localization in the image. CNN with a final Global average pooling (GAP) layer was constructed and the weights of the GAP layer were employed to compute heat maps showing the localization of objects in an image[35]. Better results were obtained with Grad-CAM++ which is a variation of CAM that uses the gradient of the output class with respect to the activation of the feature maps to find better saliency maps[36]. Roy et al. proposed a task aware selection of features by learning a DNN for action recognition using video as input. They extract a 426 trajectory and motion features, and after learning the DNN, they study the activation potential normalized over all layers to quantify feature relevance. The authors employ the first layer activation functions to define a contribution measure for each feature [37]. Feature relevance was also studied during Parkinson disease using data mining techniques [38]. Adversarial representation learning methods were employed for robust general seizure detection models. In one study, a deep CNN model that employs 2-second window raw data as input was analyzed where the authors employ the weights of the learned model to visualize internal functions of the network and extract feature maps. Using the maps as receptive fields in the intermediate layers, they investigate domain-specific knowledge and class-discriminative features using correlation maps in different frequency bands which were further processed to construct scalp topographies [29]. None of these methods can provide deeper insights about the data. Our proposed method is designed to keep track of the used features and provide explainability of the measures used. The features that trigger the decision of the model will be revealed and this will provide valuable information that can be related to the type of disease or type of interactions during seizure and non-seizure episodes. We will also show that different seizure patients exhibit diverse feature relevance maps which can be used for further analysis.

### III. Proposed Methodology

#### A. Overview of the method

As depicted in figure 1, epileptic EEG data is segmented into short-duration fixed length windows from which we extract the five common brain rhythms δ (2−4 Hz), θ (4−8 Hz), α (8−13 Hz), β (13−30 Hz), γ (greater than 30Hz) using Butterworth bandpass filters. For each of these rhythms, we compute seven connectivity measures: Spectral matrix, inverse of spectral matrix, coherence, partial coherence, directed coherence, partial directed coherence, and phase-locking value. These connectivity measures are arranged in a tensor and fed as features to a deep seizure detection network. We evaluate and compare four different deep learning models by manipulating the fusion mechanism. Since our features represent different perspective connectivity measures, we intend to benefit from the rich information found in these features. Similar to [37], we use the activation values of the learned model to output the type of connectivity that quantifies participation percentage in making the model make a certain decision.

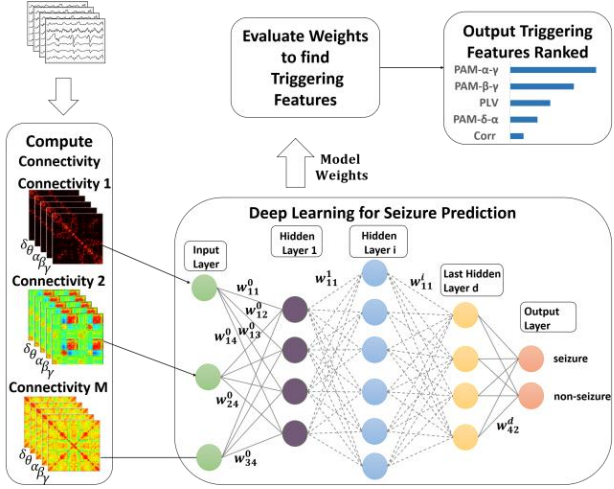

Fig. 1. The overall Workflow

### B. Data Preparation and preprocessing

The raw data were first preprocessed to extract 20-seconds long windows of seizure and non-seizure data which were manually extracted using the labels provided with the data. Each of these signals is divided further into ten 2-seconds long sub-windows resulting in a sequence of 10 seizure sub-windows that are fed to bandpass filters to extract five rhythms for each.

### C. Feature Extraction

Taking our channels as a multivariate process $Y(n)$, the multivariate linear shift-invariant filter representation can be expressed by:
$$Y(n) = \sum_{k=-\infty}^{\infty} H(k)U(n-k) \quad (1)$$

Where $U(n)$ is a vector of $m$ zero mean inputs and $H(k)$ is $m \times m$ matrix representing a filter impulse response.

On the other hand, the multivariate autoregressive (MVAR) model of order $p$ can be expressed as:
$$Y(n) = \sum_{k=1}^{p} A(k)Y(n-k) + U(n) \quad (2)$$

where $U(n)$ can be considered as uncorrelated zero mean Gaussian noise. This model can allow defining interactions between different signals such as coupling and causality using the matrices $A(k)$ since the term $a_{ij}(k)$ quantifies the causal linear interactions between $y_i$ and $y_j$ at lag k.

In the frequency domain, using the Fourier transform, the above equations yield $Y(f) = H(f)U(f)$ and $Y(f) = A(f)Y(f) + U(f)$. By comparing the two spectral representations above, one can derive the following relation: $H(f) = [I - A(f)]^{-1} = \bar{A}(f)^{-1}$.

The cross-spectral density matrix S(f) and its inverse P(f) are defined by $S(f) = H(f) \Sigma H^H(f)$ and $P(f) = \bar{A}^H(f) \Sigma^{-1} \bar{A}(f)$ where the superscript $H$ represents the Hermitian transpose and $\Sigma$ represents the covariance of $U(n)$. The coherence between the two signals $y_i$ and $y_j$ at a frequency f is can now be derived as:

$$YCoh_{ij}(f) = \frac{h_i(f) \Sigma h_j^H(f)}{\sqrt{h_i(f) \Sigma h_i^H(f)} \sqrt{h_j(f) \Sigma h_j^H(f)}} \quad (3)$$

while the directed coherence can be expressed by:
$$DC_{ij}(f) = \frac{\sigma_j H_{ij}(f)}{\sqrt{\sum_{m=1}^{M} \sigma_m^2 |H_{im}(f)|^2}} \quad (4)$$

where $\sigma_i^2$ represents the variance of signal $u_i$.

The partial coherence can also be derived in a similar fashion and can be represented by:
$$PCoh_{ij}(f) = \frac{\bar{a}_i^H(f) \Sigma^{-1} \bar{a}_j(f)}{\sqrt{\bar{a}_i^H(f) \Sigma^{-1} \bar{a}_i(f)} \sqrt{\bar{a}_j^H(f) \Sigma^{-1} \bar{a}_j(f)}} \quad (5)$$

from which we can infer the partial directed coherence, PDC, defined by:

$$PDC_{ij}(f) = \frac{\frac{1}{\sigma_j} \bar{A}_{ij}(f)}{\sqrt{\sum_{m=1}^{M} \frac{1}{\sigma_m^2} |\bar{A}_{im}(f)|^2}} \quad (6)$$

On the other hand, phase locking value is a measure that quantifies the synchrony between two signals. The signals $y_i$ and $y_j$ are first band pass filtered in the specified frequency bands, and then Hilbert transform is applied to extract the corresponding phases $\phi_i$ and $\phi_j$. The phase locking value (PLV) can be expressed as:

$$PLV_t = \frac{1}{N} \left| \sum_{n=1}^{N} e^{j[\phi_i(t,n) - \phi_j(t,n)]} \right| \quad (7)$$

where N is the number of samples considered per window.

### D. Deep Learning Model Architectures

The data formed is in the form of a tensor of dimension $7 \times 10 \times 19 \times 19 \times 5$. Each data sample takes in 10-time windows of the 7 connectivity features of 2-second intervals, each window is represented by a 19x19x5 matrix where the third dimension represents the frequency band and 19x19 is the connectivity matrix. Each row and column represent an EEG channel so one entry in this matrix represents the connectivity of the channel along the row with the channel along the column in one of the 5 different frequency bands. In our four architectures, all convolution operations use one filter because we are trying to capture a numerical combination of the input matrix and not a characteristic (i.e. shade or edges, etc..) since it is not an image. Furthermore, they have a similar base architecture, but they use different schemes to combine the different features. For our first architecture shown in figure 2, we separate the 7 features and the five frequency bands of each feature, resulting into 35 independent inputs, and feed them into separate, identical blocks as depicted in figure 2A. We process each input as a time series so each window passes through one 2D convolution layer of kernel size of (19,1) .

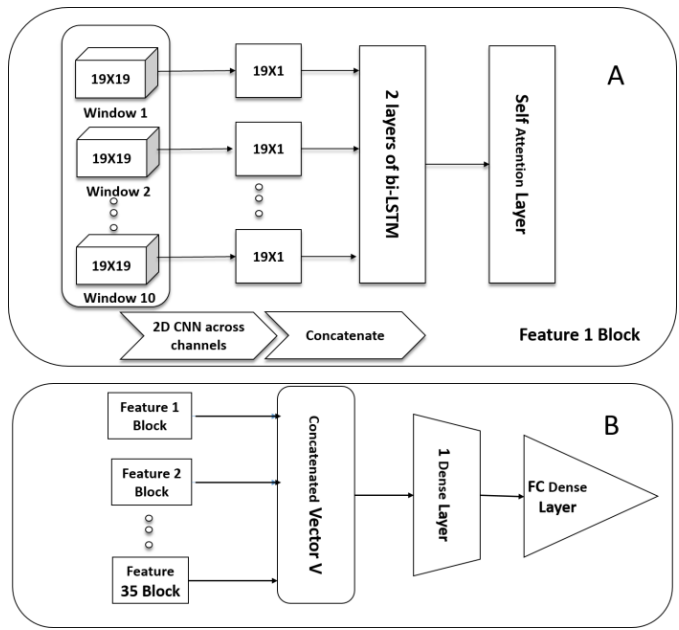

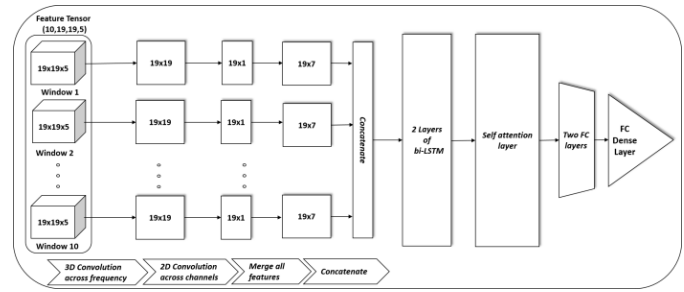

Fig. 4. Model 3 architecture

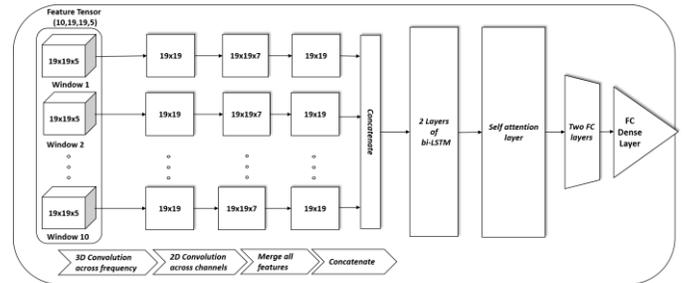

Fig. 5. Model 4 architecture

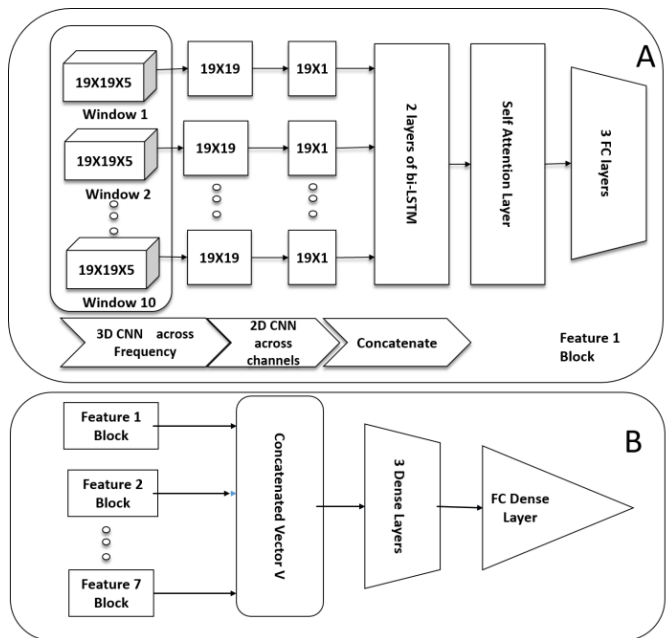

Fig. 2.

Fig. 3.

This operation helps us to condense all the relations that a channel has with the other into one number. Since we obtained one feature vector per window, we feed them to an LSTM block followed by a self-attention layer.

Each one of the 35 inputs pass through this block and at the end, the vectors obtained from the last FC layers are concatenated and fed to an FC layer for classification as shown in figure 2B. This fusion scheme assumes total independence of the features and frequency bands, the feature vectors are concatenated before the last FC layer. The second fusion scheme, instead of separating the 7 features and their 5 frequency bands, we only separate the features. this model assumes independence of the features but not the frequency bands of the individual features. We apply a 3D convolution of kernel size (1,1,5) on the windows across the frequency axis to capture the relationship between the frequencies of an individual feature. The subsequent operations are similar in nature to those in model 1 and can be seen in figure 3A and 3B. For the third fusion scheme, after having a feature vector of size 19 for each window of each feature we concatenate the feature vectors of similar time steps together, obtaining a 19x7 matrix then we proceed with a 2D convolution layer of kernel size (1,19) to get one feature vector of size 19.

We obtain channel-wise combination of the different features which we feed into an LSTM block followed by an attention layer and FC layers (figure 4). Figure 5 depicts the fourth fusion scheme, after having obtained a 19x19 matrix for each window of each feature we concatenate the feature matrices of similar time steps together, obtaining a 19x19x7 matrix for each time step then we proceed with a 3D convolution layer of kernel size (1,1,7) to get a 19x19 matrix in order to get a frequency wise combination of the different features and then proceed with a 2D convolution layer of kernel size (1,19) followed by an LSTM block, an Attention layer, and FC layers. Thus, our 4 fusion schemes aim at deducing the level at which relationships between the features are the strongest. The first scheme assumes no relation between features and frequencies, the second assumes a high-level relation between the features and strong relation between the frequencies, the third assumes a channel-wise relation for the different features and the fourth one assumes a frequency wise relation for the different features.

*E. Feature Relevance and Impact on Decision*

For the XAI part, we planned to address one of the fundamental neuroscientific questions concerned in finding relevant temporal and spatial scales necessary for given behavior [28] by doing a statistical study for the relationships between the derived connectivity measures varying between spectral, causal, and phase-related, that characterize the brain signals on one hand and the seizure-nonseizure for patient-specific and cross-patient cases on the other hand. We aim at finding some link between the explained features contribution results to some of the scientific facts concerning the seizure detection and neurology fields.

To apply XAI, we first targeted our research on the preprocessing and extracting different connectivity measures (SM, ISM, DC, C, PDC, PC, PLV) across different frequency bands $(\delta, \theta, \alpha, \beta, \gamma)$ that we believe, based on some prior knowledge and experimental studies, have a direct impact on seizure detection [13][25][26][27]. Then, we tried to achieve our explainability in the post-modeling stage using the input-based explanation drivers methods where we base our feature study on the output predictions [42]. As we see in the structure of model 2, we tried separating each feature's CNN-LSTM unit alone and concatenate them at a later stage, in such a way we can simply do our study using the concatenation and the first dense layer only. This allows us to keep track of the features. The concatenation layer will combine the outputs from the 7 features separated CNN-LSTM units into one flatten layer as depicted in figure 5 as vector $V$. Our further investigation was based on the feature extraction paper [37] at the first dense layer. To get the relevance of each of the input neurons of the flatten layer, we first calculated the average absolute activation potential $p_{ij}$ contributed by the $i^{th}$ dimension of the input:

$$p_{ij} = \frac{1}{M}\sum_{K=1}^{M}|a_{ij}^{k}| \quad (8)$$

where the activation $a_{ij}$ is: $a_{ij} = x_i w_{ij} + b_j$.

Then, we find the relative contribution $c_{ij}$ of the $i^{th}$ input dimension towards the activation of the $j^{th}$ hidden neuron:

$$c_{ij} = \frac{a_{ij}}{\sum_{i=1}^{7xN_1} p_{ij}} \quad (9)$$

In order to get the total net contribution is of an input dimension $i$ overall hidden layers, we then computed the net $c_i^+$ of all $c_{ij}$'s for every input $i$ overall $j$:

$$c_i^+ = \sum_{j=1}^{N_2} Relu(c_{ij}) \quad (10)$$

Since, our input is a set of neurons per feature in order of the sequence SM, ISM, DC, C, GDPC, PC, PLV respectively, then we further summed each of that sets $c_i^+$ alone to get the net $c^+$ per feature. According to the feature extraction paper, the higher the $c_i^+$ contribution of an input dimension, the more likely it is its participation in hidden neuronal activity and consequently, classification [37].

To begin our study, we first input the whole preprocessed dataset, cross-patient study, for seizure and non-seizure cases separately in the first part of the model, the layers before the concatenation part, and extracted the embeddings or the 400 neuron per feature and then got our results according to the technique described above. The analysis is further extended to find inter-patient feature relevance variations.

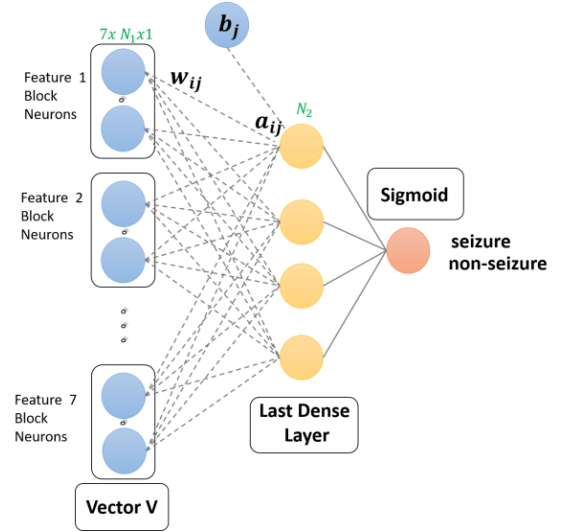

Fig. 6 Last layer architecture weights and activations from which relevance of features are computed

## IV. EXPERIMENTAL RESULTS

*A. Experimental Setup*

CHB-MIT is an EEG dataset collected at Children's Hospital in Boston. It has 24 cases of epilepsy. EEG was acquired using the international 10-20 system sampled at 256 samples per second with a 16-bit resolution. Overall, 198 seizures are annotated with the beginning and the end of seizures. This particular dataset was chosen because it was used in many state of the art papers which allows us to compare our deep learning model's performance. As shown in figure 6, we extract from this dataset 20 second intervals from seizure and non-seizure episodes. The seizure of length less than 20 seconds, which were very few, were ignored in this study and the seizure episodes of duration greater than 20 seconds were dissected into 20 second intervals. The remainders were also considered by taking 20 seconds from the end. As for the non-seizure episodes, four 20-second intervals were taken randomly from every dataset. The total number of 20 episode intervals collected contained 543 seizure intervals and 801 non-seizure intervals.

*B. Validation Metrics*

To evaluate the performance of the models, we use the statistical measures of binary classification. We denote the seizure label as the positive class and the non-seizure case as the negative class. We first define the following terms by

- True Positive (TP) : number of hits, correctly classified positives
- False Positive (FP): number false alarms, classified as seizure while it actually has no seizure.

- False Negative (FN): number of misses, classified as non-seizure while it actually is seizure.
- True Negative (TN): number of correct rejections, correctly classified negatives.

Sensitivity measures the percentage of positive class members that are correctly identified and is given by:

$$Sensitivity = \frac{TP}{TP + FN}$$

Specificity gives the percentage of negative class members that are correctly identified whose formula is:

$$Specificity = \frac{TN}{TN + FP}$$

Precision finds the positive predictive rate which explains how much of the positives were identified.

$$Precision = \frac{TP}{TP + FP}$$

Finally, the accuracy is computed using:

$$Accuracy = \frac{TP + TN}{TP + TN + FP + FN}$$

C. Feature Extraction

Each of these 20-second intervals was first filtered in the five frequency bands described above and further dissected into ten 2-second intervals each of which is processed to extract the seven described features. The ten sub-intervals are then gathered into a single tensor resulting in a tensor of size $7x10x19x19x5$ as shown in figure 7.

D. Performance Results of the proposed models

The models were evaluated on a data split of 85% and 15%. After fine tuning, the first model performed the best with a 97.03% accuracy as you can see in Table 2. The results also show that the earlier the fusion scheme the weaker the relation between the features. Since our data dissection method is totally new, no quantitative comparative analysis was performed. We train all our models twice, the first time non-seizure data has a label of 1 and seizure data has a label of 0 and the second time, using the resulting weights from the first time, the labels are flipped, that procedure forces the model to learn more robust features for each class. The data was too big to load into memory, so we had to write our own data generator that fetches batches of 32 samples from the folder. The hyper parameters for each model are as follows:

- Model1: a relu activation function for both dense and convolution layers, 2 LSTM layers and 1 dense layers of 100 neurons for each block and a dropout rate of 0.5. The optimizer used is Adam and 30 epochs for both training phases.
- Model2: a relu activation function for both dense and convolution layers with a Spatial Dropout2D of 0.07, we use 2 LSTM layers and all dense layers have Dropout of 0.5. We use 2 dense layers of 263 and 20 neurons respectively. The optimizer used for the first training phase was RMSprop for 17 epochs, after the flip Nadam was used as optimizer for 8 epochs.
- Model3: a relu activation function for both dense and convolution layers, 2 LSTM layers and 3 dense layers of 500 neurons for each block and a dropout rate of 0.5 and l2 regularization. The optimizer used is Adam and 80 epochs for both training phases.
- Model4: a relu activation function for both dense and convolution layers, 2 LSTM layers and 3 dense layers of 200 neurons for each block and a dropout rate of 0.3 and l2 regularization. The optimizer used is Adam and 100 epochs for both training phases.

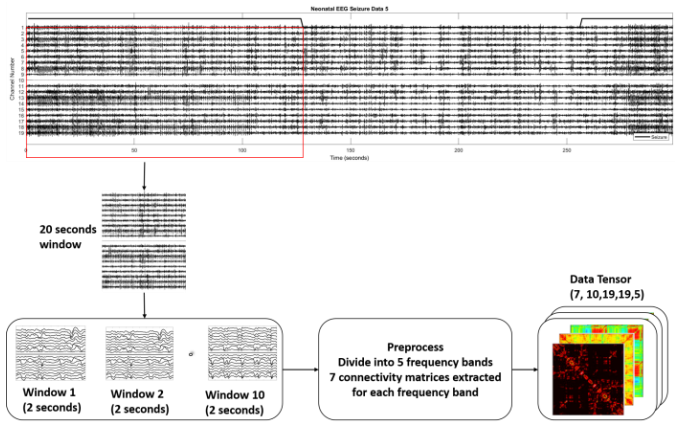

Fig. 7. Data preparation and feature extraction

The performance of the different modes on training and testing data are tabulated in table II. The results are the average of ten runs with different splits each time. The overall accuracy, sensitivity (), specificity (), and the precision () are shown. While the model is well fit to the training data in all models, unseen testing data performance is considered to choose the best model. Model 1 performs the best in all measures recording a sensitivity of 97.65%, a specificity of 96.58%, precision 95.4%, and an overall accuracy of 97.03%.

Table II. Performance Metrics Across all proposed models

| | Data | Sensitivity | Specificity | Precision | Accuracy |
|---|---|---|---|---|---|
| **Model 1** | Training | 100.00 | 99.85 | 99.78 | 99.91 |
| **Model 1** | Testing | **97.65** | **96.58** | **95.40** | **97.03** |
| **Model 2** | Training | 99.13 | 98.10 | 97.22 | 98.51 |
| **Model 2** | Testing | 94.67 | 96.06 | 93.42 | 95.54 |
| **Model 3** | Training | 99.79 | 100.00 | 100.00 | 99.91 |
| **Model 3** | Testing | 80.23 | 92.24 | 88.46 | 87.13 |
| **Model 4** | Training | 96.08 | 96.49 | 94.84 | 96.32 |
| **Model 4** | Testing | 77.22 | 86.18 | 78.21 | 82.67 |

E. Feature Relevance Results

Figure 8 represents the overall feature relevance of all the data. We can see that spectral matrix and partial coherence have higher relevance during seizures on average while non-seizure

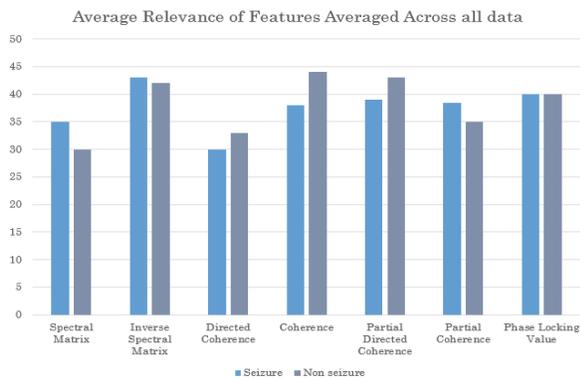

Fig. 8. Feature relevance cross all data

decisions are made more effective with the coherence and the partial directed coherence.

Cross-patient test results show the variation of the feature relevance diagrams for each of the tests as can be seen in figures 9 and 10. We notice that every patient had different feature relevance plot than the other, such that there is a different way in assigning the relevance and weights of each feature which can be interpreted as a validation of the scientific fact that EEG patterns in seizure patients are highly variable across patients. On the other hand, there were some changes across the same patient results which can be explained on basing the features on both time and frequency domains and EEG seizure pattern data are highly dynamic in nature for the same patient.

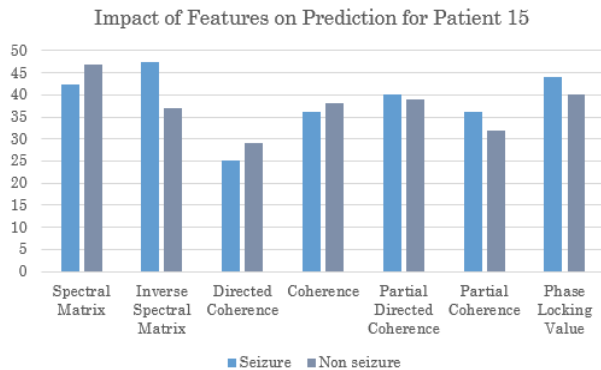

Fig. 9. Feature relevance for patient 15

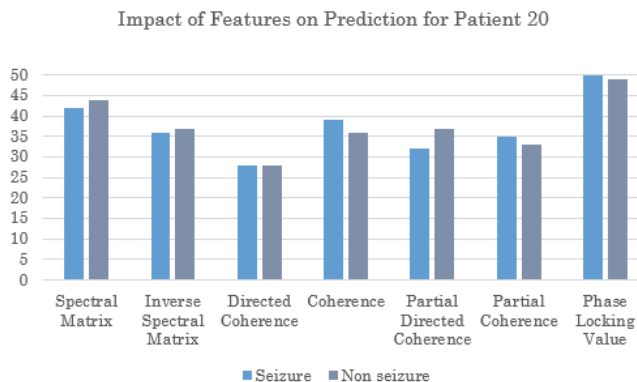

Fig. 10. Feature relevance for patient 20

Another interesting finding is that the directed coherence feature is often assigned with the least net contribution $c^+$. The phase-locking value is often assigned with good relevance value in seizure and non-seizure compared to other features which might be an indication for the fact that PLV is one of the undirected connectivity measures that can infer the disconnection of the parts of the brain where the seizure onset zones get disconnected [26].

## V. DISCUSSION AND FUTURE WORK

EEG data analysis has been studied using EEG data and many of the techniques rely on the data dissection and the features used. In this study, we adopted a sequential portion of EEG data which contains information about a sequence of ten consecutive EEG sub-windows. This choice was made because during seizures, the signals may alternate from seizure to non-seizure states. Also, during non-seizure data, short bursts of seizure activity may arise and can be labeled as seizure if taken separately. The BiLstm structure between these ten windows can learn relationships between the windows during a seizure and avoid such misclassifications. Besides, seizure data is generally noisy and stochastic in nature, and having relating higher-level information between consecutive windows can help learn more complex relationships across time.

Our study focuses on discriminating seizure/non-seizure while providing explanations of our learned model which make it more presented different CNN-Lstm models for detecting seizure based on long windows. Our models used different fusion strategies where the first two models combine the features at the decision level, and the two combined them at the input level. We were able to show that combining features at the decision layers yield in a much better performance which can be an indication that the features are better learned when separated in the feature extraction part. In our study, we made use of the latest techniques and advancement mainly in the regularization and the normalization methods that helped our architecture in achieving better results. We employed various fusion mechanisms and conclude that fusion at the input is not performing well compared to fusion the features at the end which makes CNN able to learn and extract its high order representation better.

Feature extraction was computationally very expensive and took few days. This can be accelerated using GPU programming and distributed and parallel computing methodologies. Many extensions are possible at the level of architecture, feature selection, and explainablity. We can investigate different architectures to capture other relations, since the assumption that we started from in this study is that the smallest input possible is a window of the same feature in different frequencies, we would have to investigate if the smallest possible input window of the same frequency in different features yields better accuracy showing that the seizure is more related at the frequency level rather than the feature level and we will be able to do XAI to deduce which frequency band is the most important. Our methods need to encompass other EEG datasets to find more general models and analyze other seizure patient biomarkers. Finally, our methods can be extended to learn sequence relationships at transition episodes where states shift from pre-ictal to ictal as well as ictal to post-ictal transitions and to characterize the state transitions during seizures.

## VI. Conclusion

Seizure detection using a sequence model was proposed in this study. We have shown that relating higher resolution data together as a sequence can characterize differences between seizure and non-seizure data. Among the four studied models based on deep learning, model 1 which used fusion at the level of the decision recorded 97.03% accuracy. The models were based on different neural network architectures mainly on CNN and LSTM with attention layers. The learned weights of the model helped understand the relevance of the chosen features and further showed that they can represent cross-patient discriminative features and open way to many future studies for seizure analysis.